%% file: flashformer.tex
\documentclass{article}

\usepackage{microtype}
\usepackage{graphicx}
\usepackage{subfigure}
\usepackage{booktabs} 

\usepackage[preprint]{mlsyspreprint}

\input{math_commands}

\usepackage[utf8]{inputenc} 
\usepackage[T1]{fontenc}    
\usepackage{hyperref}       
\usepackage{url}            
\usepackage{booktabs}       
\usepackage{amsfonts}       
\usepackage{nicefrac}       
\usepackage{xcolor}         

\usepackage{subcaption}
\usepackage{cleveref}
\usepackage{xspace}
\usepackage{wrapfig}
\usepackage{array}
\usepackage{listings}
\usepackage[font=small,labelfont=bf]{caption}

\definecolor{darkblue}{rgb}{0, 0, 0.5}
\hypersetup{colorlinks=true, citecolor=darkblue, linkcolor=darkblue, urlcolor=darkblue}

\newcommand{\flashformer}[0]{\textsc{FlashFormer}\xspace}

\mlsystitlerunning{FlashFormer: Whole-Model Kernels for Efficient Low-Batch Inference}

\begin{document}

\twocolumn[
\mlsystitle{\textsc{FlashFormer}: Whole-Model Kernels for \\ Efficient Low-Batch Inference}

\mlsyssetsymbol{equal}{*}

\begin{mlsysauthorlist}
\mlsysauthor{Aniruddha Nrusimha}{mit}
\mlsysauthor{William Brandon}{mit}
\mlsysauthor{Mayank Mishra}{ibm}
\mlsysauthor{Yikang Shen}{ibm}
\mlsysauthor{Rameswar Panda}{ibm}
\mlsysauthor{Jonathan Ragan-Kelley}{mit}
\mlsysauthor{Yoon Kim}{mit}
\end{mlsysauthorlist}

\mlsysaffiliation{mit}{Massachusetts Institute of Technology, Cambridge, MA, USA}
\mlsysaffiliation{ibm}{MIT-IBM Watson AI Lab, Cambridge, MA, USA}

\mlsyscorrespondingauthor{Aniruddha Nrusimha}{anin@mit.edu}

\mlsyskeywords{Machine Learning, Large Language Models, Kernel Optimization, Low-Batch Inference}

\vskip 0.3in

\begin{abstract}
The size and compute characteristics of modern large language models have led to an increased interest in developing specialized kernels tailored for particular training and inference workloads. Existing kernels primarily optimize for compute utilization, targeting the large-batch training and inference settings. However, low-batch inference, where memory bandwidth and kernel launch overheads are significant factors, remains important for many applications of interest such as in edge deployment and latency-sensitive applications. 
This paper describes \flashformer, which fuses the entire transformer forward pass into a single kernel for  accelerating low-batch inference of large language models.
Across various model sizes and quantizations settings, \flashformer achieves  nontrivial speedups compared to existing  inference kernels.
\end{abstract}
]

\printAffiliationsAndNotice{\mlsysEqualContribution}

\input{sections/introduction}

\input{sections/background}

\input{sections/motivating_study}
\input{sections/method}

\input{sections/results}
\input{sections/RW}
\input{sections/conclusion}


\bibliography{flashformer}

\newpage
\appendix

\input{appendices/algorithims}

\end{document}

%% file: math_commands.tex
\usepackage{amsmath,amsfonts,bm}

\def\eqref#1{equation~\ref{#1}}

\def\1{\bm{1}}

\def\va{{\bm{a}}}

\def\vg{{\bm{g}}}

\def\vx{{\bm{x}}}

\def\mK{{\bm{K}}}

\def\mO{{\bm{O}}}
\def\mP{{\bm{P}}}
\def\mQ{{\bm{Q}}}

\def\mS{{\bm{S}}}

\def\mV{{\bm{V}}}
\def\mW{{\bm{W}}}

\DeclareMathAlphabet{\mathsfit}{\encodingdefault}{\sfdefault}{m}{sl}
\SetMathAlphabet{\mathsfit}{bold}{\encodingdefault}{\sfdefault}{bx}{n}



%% file: sections/introduction.tex
\section{Introduction}

As the utility and popularity of deep learning models has steadily increased, there has been an increasing interest in low-level \textit{kernel design}, i.e., developing specialized programs tailored to optimize particular workloads on the target hardware.
This has been particularly true for large language models (LLMs) based on the transformer architecture, due to a dramatic increase in task performance across domains for these models obtained from  \textit{model scaling} (i.e., increasing parameter counts and computational costs per forward pass) and \textit{inference scaling} (i.e., generating more tokens in parallel or serially).

While most accelerator vendors develop high performance kernels for common linear algebra routines such as matrix multiplications, these general-purpose kernels are often not sufficient for LLM inference. 
Consequently, there has been much recent work on kernels that accelerate the inference of transformer-based LLMs, such as \textsc{FlashDecoding} \citep{flashdecoding} and \textsc{PagedAttention} \citep{10.1145/3600006.3613165}. These kernels achieve efficiency gains by optimizing parts of operations common to the transformer inference pipeline. For example, \textsc{FlashDecoding} modifies the work partitioning in \textsc{FlashAttention} \citep{dao2022flashattention,dao2023flashattention2} for better parallelization during transformer decoding, while \textsc{PagedAttention} saves GPU memory by organizing the KV cache, thus allowing larger batch sizes.

\begin{figure*}[h]
    \label{fig:Pipelining}
    \centering
            \vspace{-2mm}
    \includegraphics[width=.9\linewidth]{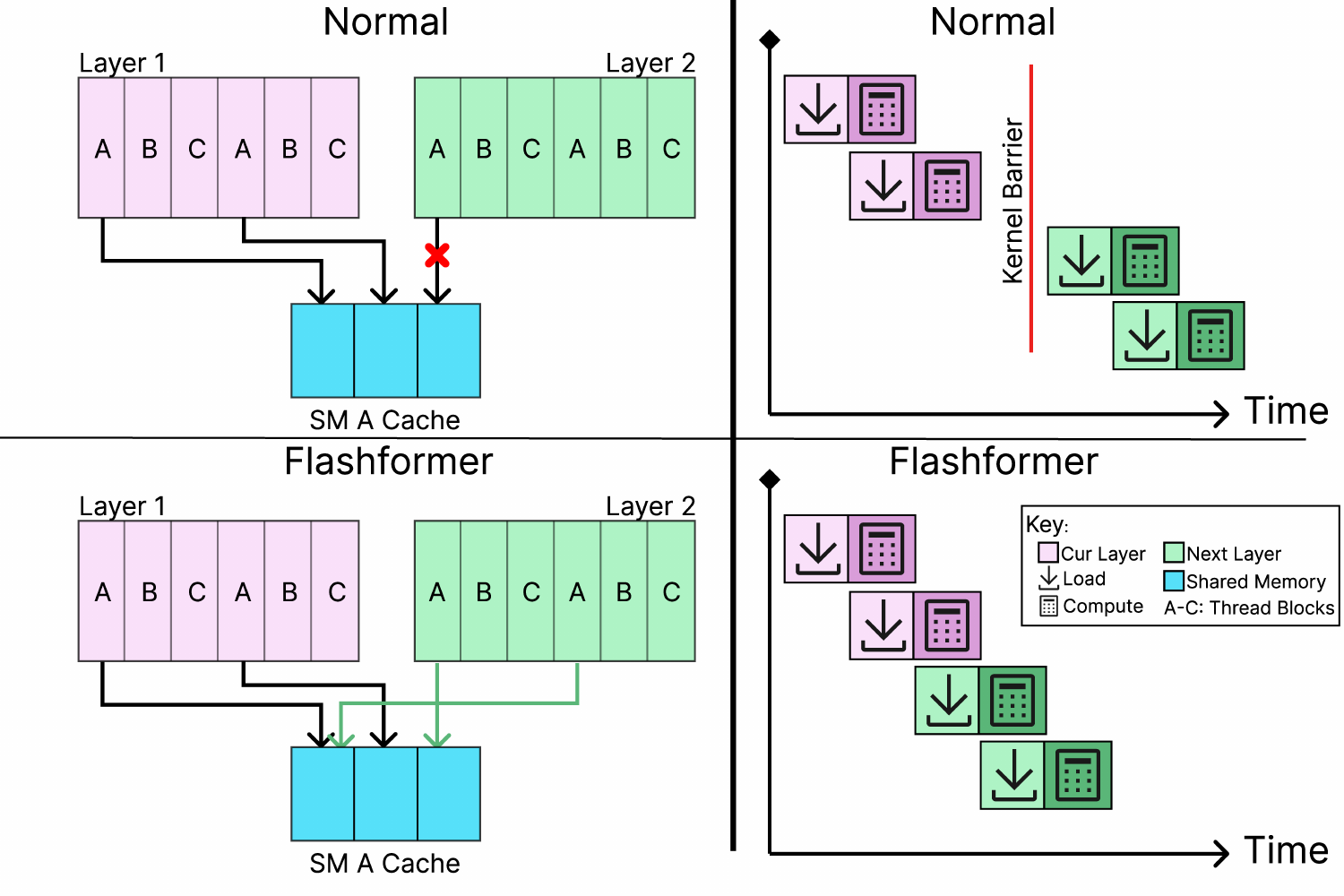}
            \vspace{-4mm}
    \caption{Depiction of the key benefit of \flashformer. The top half depicts normal transformer execution with multiple kernels. After all chunks of Layer 1 have been loaded into the SM cache, we cannot start loading Layer 2 (top left). This, coupled with kernel overhead, leads to a break in memory loads and computation (top right). With \flashformer, we fuse kernels to enable overlapping across prior kernel barriers (bottom left). This leads to more efficient overlapping and faster runtime (bottom right).}
            \vspace{-4mm}
\end{figure*}
Most existing kernels are optimized to maximize throughput for large matrix-matrix multiplications (matmuls). The key to doing so on modern accelerators is to maximize the utilization of special matrix-multiply units (MMU)---called Tensor Cores on NVIDIA hardware and MXUs on TPUs---which provide most of the computational performance of the chip.
In either large-batch inference or training, this design choice optimizes the primary bottleneck of these kernels, i.e., the computational cost of matmuls with high arithmetic intensity (computation per element of data movement).

However, many use cases operate under different constraints. For example, low-batch inference is primarily bottlenecked by the memory bandwidth costs of parameter and KV-cache movement; depending on the size of the model, kernel launch overheads that arise from launching multiple separate kernels for each transformer layer can also be a bottleneck. Low-batch inference workloads arise in many real-world applications, such as edge deployment, latency-sensitive applications, and more recently, inference-time scaling via longer rollouts. Moreover, low-batch inference does not obtain significant benefits from using MMUs on the hardware since the problem sizes are not bottlenecked by compute.

Writing fast kernels  for these workloads requires further innovation along two common directions in kernel development research: 1) Increased operator fusion, or packing an increasing number of instructions and operations into a single kernel and 2) Increased specialization, or generating kernels that more closely optimized for the particular problem.
This paper describes 
a strategy for writing a highly optimized inference kernel that runs the entire forward pass of a transformer-based LLM in a single kernel, and is specialized for a particular model configuration (such as the number of attention heads, embedding dimension, etc) and the target hardware architecture.
We use meta-programming, a shared pipelined buffer, and fast synchronization mechanisms to create a fully fused and specialized kernel. The resulting kernel, which we call \flashformer, exploits increased overlapping of memory movement and the single fused kernel amortizes the launch overhead for the kernel over the entire forward pass. We apply this kernel to accelerate low-batch inference of Llama models \citep{grattafiori2024llama}, and find that it offers consistent speedups over existing decoding kernels.

%% file: sections/background.tex
\vspace{-1mm}
\section{Background}
\vspace{-1mm}
\subsection{Transformer Architecture}
\vspace{-1mm}

We target the Llama \citep{grattafiori2024llama} family of transformer architectures, which make use of the  pre-RMSNorm \citep{jiang2023pre}, Grouped Query Attention \citep{ainslie2023gqa}, RoPE \citep{su_roformer_2021}, and SwiGLU FFN layers \citep{shazeer_glu_2020}. 
Each layer of a transformer model consists of six matrix operations, commonly grouped into five sublayers. Each sublayer has one matrix operation, except for the ``core attention'' layer,\footnote{Since ``attention layer'' could be ambiguous,  we use ``core attention'' to refer to the operation $\operatorname{softmax}(\alpha \mQ\mK^\top)\mV$.} which has two. 
The matrix operations per layer with their respective input and output activations are listed in \Cref{tab:sublayers}. These components are common to many other recent families of LLMs.

During autoregressive token-by-token generation, each matrix operation takes as input a large matrix stored in global memory as well as an activation computed by a previous operation.

Transformer LLMs are parameterized by the number of layers, model dimension, FFN dimension, head dimension, and the number of attention and key-value heads. The models we generate kernels for  are shown in \Cref{tab:llama configs}.
\begin{table*}[t]
    \centering
    \begin{tabular}{llll}
        \toprule
        \textbf{Matrix Operation} & \textbf{Matrix Load} & \textbf{Input} & \textbf{Output} \\ 
        \midrule
        $\bm{QKV}$ Projection &  $\mW_{\bm{QKV}}$ & $\text{RMSNorm}(\vx)$ & $\mQ[i],\mK[i],\mV[i]$ \\
        $\mQ \mK^\top$ computation & $\mK[\hspace{0.3mm}:\hspace{-0.7mm}i]$ & $\mQ[i]$ & $\mS[i]$ \\
        $\bm{PV}$ computation & $\mV[\hspace{0.3mm}:\hspace{-0.7mm}i]$ & $\text{softmax}(\alpha \mS[i])$ & $\mO$ \\
        Attention Output Projection & $\mW_{\va_{out}}$ & $\text{vector}(\mO)$ & $\va_{out}$ \\
        GLU In/Gate Projection & $\mW_{FFN1}$ & $\text{RMSNorm}(\vx + \va_{out})$ & $\vg_{inter}$ \\ 
        GLU Output Projection & $\mW_{FFN2}$ & $\text{SwiGLU}(\vg_{inter})$ & $\vg_{out}$ \\
        \bottomrule
    \end{tabular} 
        \vspace{-2mm}
    \caption{Transformer matrix operations for decoding for Llama 3.1. Here $i$ is the current sequence position, and $\vx$ is the residual stream.
    $\mW_{\bm{QKV}}, \mW_{\va_{out}} \mW_{FFN1}$ and $\mW_{FFN2}$ represent corresponding parameter matrices, and $\mQ, \mK$ and $\mV$ represent the query, key, and value matrices respectively. $\mS$ represents the attention logits, $\mP$ the attention weights, and $\va_{out}$ the output of attention. $\vg_{inter}$ and $\vg_{out}$ represent the output of the GLU input/gate projection and the final GLU output of the GLU layer respectively.}
    \label{tab:sublayers}
\end{table*}
\begin{table*}[t]
    \centering

    \begin{tabular}{lllllll}
        \toprule
        \textbf{Model} &  \textbf{Layers} & $d_{model}$ & $d_{inter}$ & $d_{head}$ &$n_{q\_heads}$ & $n_{kv\_heads}$\\
        \midrule
        Llama 3.1 8B & 32 & 4096 & 14336 & 128 & 32 & 8\\
        Llama 3.1 70B & 80 & 8192 & 28672 & 128 & 64 & 8\\
        \bottomrule
    \end{tabular}
        \vspace{-2mm}
    \caption{Configuration of Llama 3.1 8B and 70B, for which we develop \flashformer kernel.}
    \label{tab:llama configs}
    \vspace{-3mm}
\end{table*}

\vspace{-1mm}
\subsection{GPU Programming and Architecture}
\vspace{-1mm}


GPU programming optimizes over three interleaved hierarchies: the program hierarchy, the hardware hierarchy, and the memory hierarchy. The program hierarchy structures computations into kernels, thread blocks, warps, and threads to effectively expose parallelism.\footnote{We use the NVIDIA GPU hardware terminology throughout this work; other architectures use different terms for similar hardware components.} The hardware hierarchy maps thread blocks onto streaming multiprocessors (SMs) which is sub-divided into multiple warp schedulers responsible for running a warp which is a collection of 32 threads  executing the same instruction. Individual threads are responsible for the core computations. Finally, the memory hierarchy manages the placement of data in registers, shared memory, and global memory to minimize latency and maximize throughput.
\Cref{tab:Hardware hierarchy} provides an overview of the three hierarchies.

At each level except the thread level, it is possible for multiple elements of the program hierarchy to exist on the same hardware if they satisfy certain hardware and memory constraints. In particular, multiple warps can exist on one warp scheduler provided there is enough space in registers, andmultiple thread blocks can co-locate on an SM if there is enough shared memory and register space.

\begin{table*}[ht]
    \centering
    \begin{tabular}{lllll}
    \toprule
         \textbf{Program Level} & \textbf{Hardware Level} & \textbf{Count} & \textbf{Memory Pool} & \textbf{Capacity} \\
    \midrule
         Kernel & GPU & 1 &Global (HBM) & 80 GB \\
         Threadblock & SM & 132 &Shared & 228KB/SM\\
         Warp & Warp Scheduler & 4/SM &-&-\\
         Thread & Cuda Core & 32/Warp &Registers& 256KB/SM, 1KB/thread\\
    \bottomrule
    \end{tabular}
            \vspace{-2mm}
            \caption{The hierarchies of the GPU and and capacities on an H100 GPU. SM stands for Streaming Multiprocessor. Note that multiple thread blocks can coexist on an SM, and multiple warps on a warp scheduler.}
                    \vspace{-4mm}
    \label{tab:Hardware hierarchy}

\end{table*}

\vspace{-1mm}
\subsection{Decoding Workload and GPU Overlapping}
\vspace{-1mm}
The general strategy for overlapping in most GPU workloads is to partition the workload at each program level such that the corresponding hardware level is oversubscribed, i.e., the hardware has more work that it can execute.
This oversubscription allows tasks to overlap: when one task is stalled awaiting resources---such as data loading from global to shared memory---other tasks continue executing arithmetic computations. Similarly, when one warp is waiting on a shared memory load, another warp can issue its computational instructions.

However, this strategy does not work well for inherently sequential workloads such as the sequential computations of transformer layers. Each matrix operation in \Cref{tab:sublayers} depends on the completion of the operations before it, and each transformer layer depends on the the completion of the previous transformer layer. As a result, in normal execution there is no opportunity for overlapping between different kernels, as shown in \Cref{fig:Pipelining}. Furthermore, inference kernel launch overheads can cause a nontrivial slowdown for fast decoding kernels, as noted by \citet{thundermla}. 

\vspace{-1mm}
\subsection{Operator Fusion}
\vspace{-1mm}
One solution to the problem of overlapping is \emph{kernel fusion}, which involves taking two or more kernels and combining them into a single kernel. Existing kernel fusions can be divided into two types.
The first type fuses a single (or multiple) pointwise operations (such as an activation function) into a matrix operation. 
These pointwise operations are directly applied to either the input of the matrix or its output or both. These types of fusions are now commonly automated through tools such as XLA \citep{50530} or PyTorch Compile \citep{Ansel_PyTorch_2_Faster_2024}. 

The second type involves restructuring multiple operations to remove the need for global synchronization and reduce redundant data movement between multiple matrix operations.
A notable recent example of this is the \textsc{FlashAttention} series of kernels \citep{dao2022flashattention,dao2023flashattention2,shah_flashattention-3_2024}, which fuses the $\mQ \mK^T$ computation, softmax calculation, and $\mP \mV$ computation into one kernel, reducing the memory movement from HBM to SRAM and significantly increasing arithmetic intensity over an unfused kernel.

%% file: sections/motivating_study.tex
\section{Preliminary Study with Deep Linear Networks}
We  motivate the development of whole-model kernels for transformer inference by demonstrating the latency benefits in the case of \emph{deep linear networks}. Insofar as deep linear networks make use of simpler primitives (i.e., just a stack of linear layers), speedups would be difficult to realize; thus, any speedups observed in this setting would provide promising motivation for fusing the entire transformer forward pass. We measure the latency reduction from eliminating kernel overheads at a variety of widths and depths, 
and ablate the latency reduction from overlapping memory movement across layers. We measure both the execution time and achieved memory bandwidth; since our workloads are memory-bound, higher memory bandwidth  indicates better hardware utilization.

\begin{figure}
    \centering
    \includegraphics[width=0.9\linewidth]{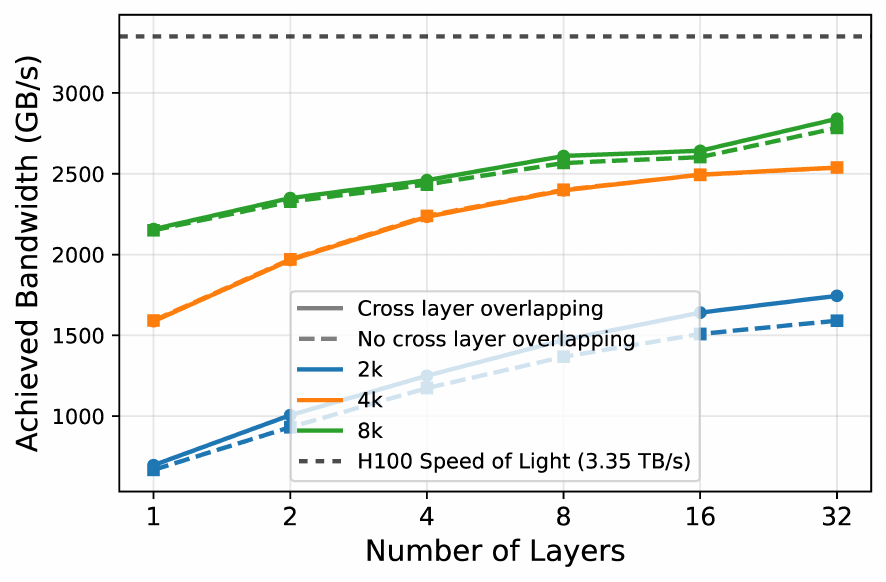}
    \vspace{-4mm}
    \caption{Memory bandwidth achieved by fused stacked linear layers. Cross layer overlapping refers to starting to load weights for future layers before computation for the current layer has finished.}
    \label{fig:multilayer_square_linear}
       \vspace{-4mm}
\end{figure}

\vspace{-2mm}
\paragraph{Kernel fusion across layers.} We first study fusing the linear layers into a single kernel thus eliminating kernel overheads. We use a cooperative square linear kernel with a dynamic number of layers with a batch-one input (i.e., a vector). As a cooperative kernel, only one thread block runs per SM. Each thread block works on a predetermined portion of the workload, and performs a global sync in between layers.  For different problem sizes (model dimension / number of layers), we perform autotuning across the number of consumer warps, number of pipeline stages, and size of each stage of the memory pipeline. The  latency is  shown in \Cref{tab:lin_layer_latency}, and the memory bandwidth is shown in \Cref{fig:multilayer_square_linear}. We find that kernel fusion across layers improves realized memory bandwidth. This effect is larger for smaller layers and diminishes as the number of fused layers increases.

\begin{table}
\centering
\begin{tabular}{l c c c c}
\toprule
\textbf{Dim} & \textbf{Layers} & \textbf{Baseline} & \textbf{Fused} & + \textbf{Overlap}  \\
\midrule
 & 1 & 0.013 & 0.013 & 0.012 \\
2K & 4 & 0.050 & 0.029 & 0.027 \\
 & 32 & 0.403 & 0.169 & 0.154 \\
 \midrule
 & 1 & 0.021 & 0.021 & 0.021 \\
4K & 4 & 0.084 & 0.060 & 0.060 \\
 & 32 & 0.675 & 0.423 & 0.423 \\
 \midrule
 & 1 & 0.062 & 0.062 & 0.062 \\
8K & 4 & 0.250 & 0.221 & 0.218 \\
 & 32 & 1.998 & 1.543 & 1.513 \\
\bottomrule
\end{tabular}
\vspace{-1mm}
\caption{Latencies of stacked linear layers, measured in ms. Baseline refers to stacking separate kernel launches, Fused combines the forward pass into a single kernel. + Overlap is same as fused but makes use of  cross layer overlapping of memory movement.}
\vspace{-4mm}
\label{tab:lin_layer_latency}
\end{table}

\vspace{-2mm}
\paragraph{Memory overlapping.} A benefit of working with a single kernel is that we can initiate memory movement for future layers before the computation for current layers finishes.  The additional benefits of memory overlapping are shown in \Cref{tab:lin_layer_latency}, representing an additional few percentage points of latency improvement over the fused kernel.

These promising initial results motivate the development of whole-model kernels for transformer inference.

%% file: sections/method.tex
\vspace{-1mm}
\section{Methods}
\label{sec:methods}

\vspace{-1mm}
\subsection{General Kernel Design Principles}
\vspace{-1mm}
\label{method-general}

We first elaborate on the key challenges in designing a family of whole-model kernels and describe how \flashformer  overcomes these challenges.

\begin{figure*}[t!]
    \centering
    \includegraphics[width=.74\linewidth]{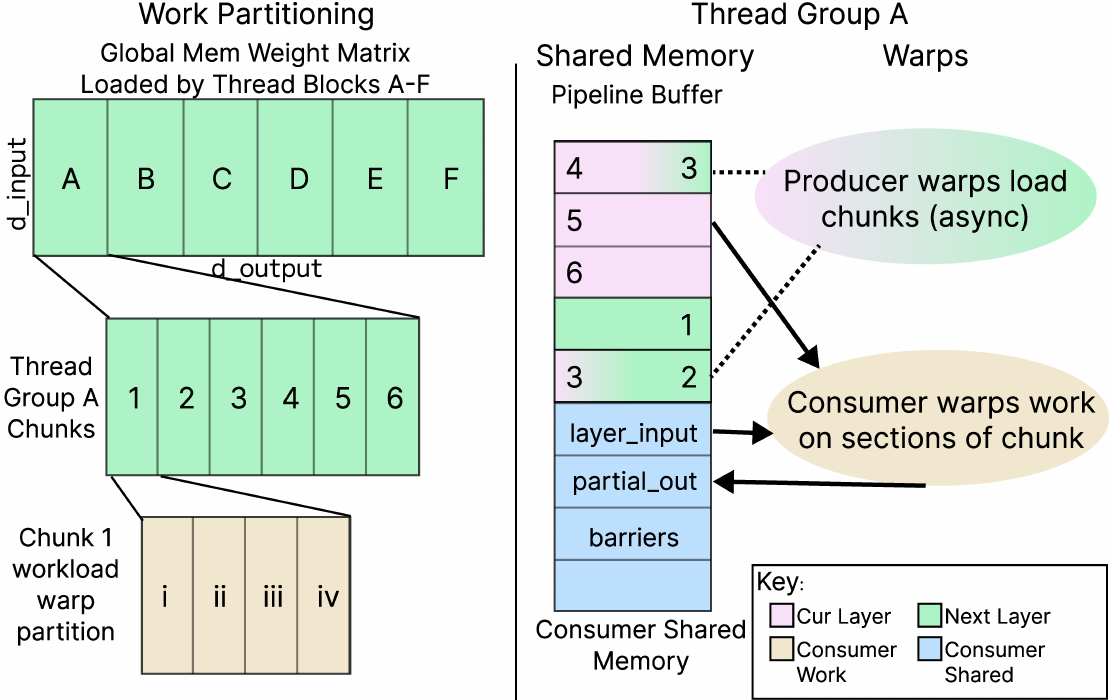}
        \vspace{-1mm}
    \caption{Left: Work partitioning between and within thread groups. 
    The work for the matrix in global memory is split between thread groups.
    The work for each thread group is split into chunks for the pipeline buffer.
    Within the chunk, the work is divided into chunks.
    Right: Thread Group A work partitioning. Producer warps manage the pipeline buffer asynchronously, while consumer warps work synchronously on sections of chunks.
    }
    \vspace{-3mm}
    \label{fig:partition}
\end{figure*}

\vspace{-2mm}

\paragraph{Metaprogramming for static optimization.} New kernels need to be implemented for each model tested. Writing such kernels manually is time intensive and prone to bugs.
Implemented kernels can vary greatly depending on factors such as model dimension and quantization settings. To address this, we developed a metaprogramming language (in Python) called \textsc{Cheetah}\footnote{\url{https://github.com/cheetah-lang}}, which, given the dimension sizes of the model, performs static workload partitioning based on predetermined rules.

\begin{figure*}[t!]
    \centering
    \includegraphics[width=.9\linewidth]{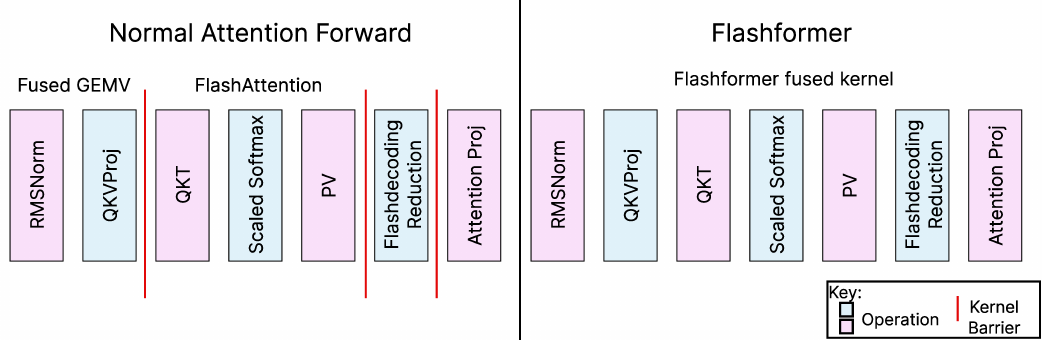}
    \caption{
    Conceptual diagram of a standard attention forward pass and our method. Left: a standard attention forward pass in an inference pipeline, broken down into four kernels. Right: FlashFormer, where all operations are part of one kernel invocation. 
    }
        \vspace{-4mm}
    \label{fig:attention}
\end{figure*}

\begin{figure}[h]
    \centering
    \includegraphics[width=1\linewidth]{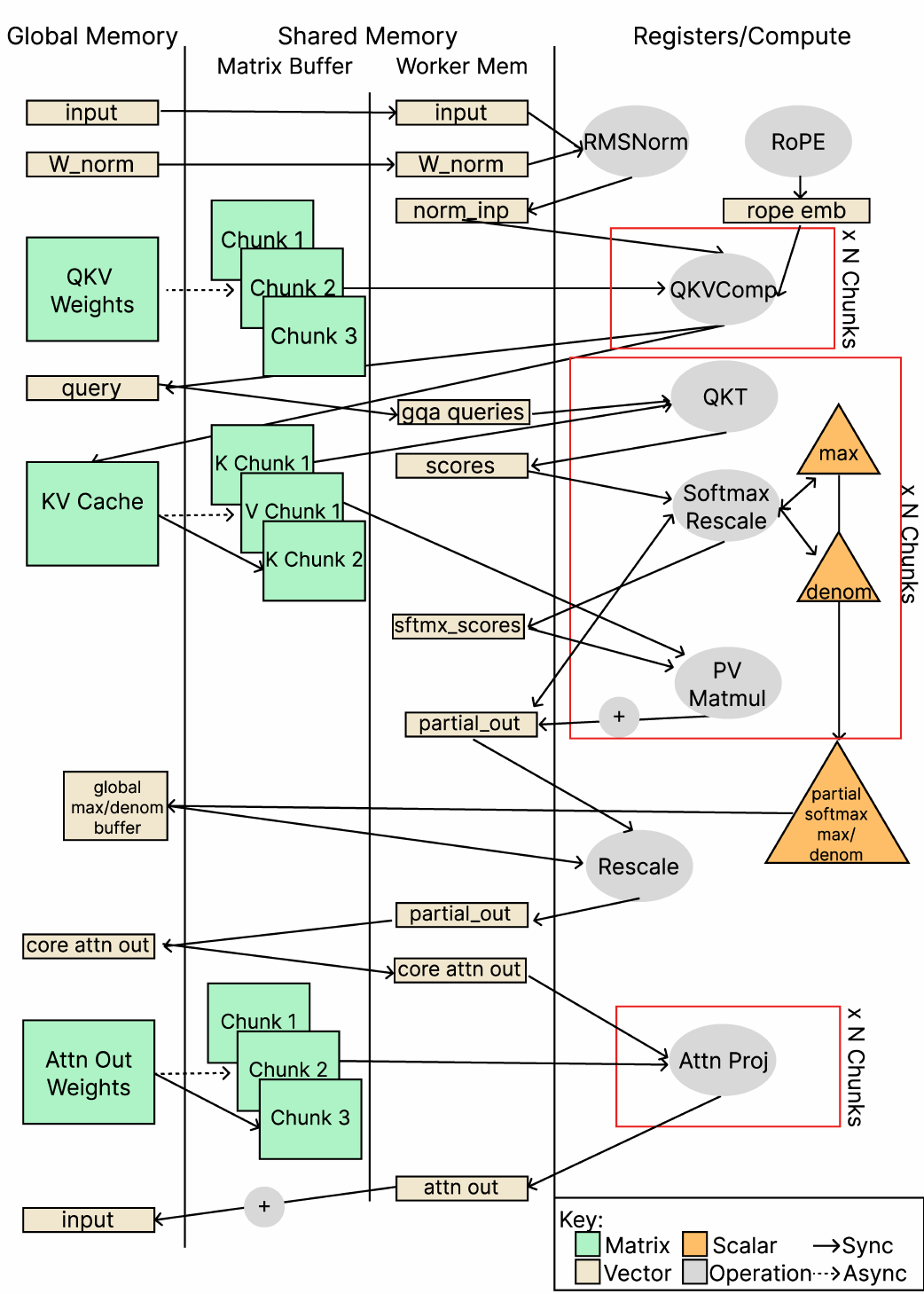}
    \caption{The Attention forward pass per layer. Large matrices (green) are loaded asynchronously from global, and do not wait on warps performing computation. The consumers works synchronously, with 3 global synchronizations per attention forward pass.
    Each of the repeated computations (in red) corresponds to a asynchronously loaded matrix split into chunks.}
    \vspace{-4mm}
    \label{fig:attn_long}
\end{figure}

Concretely, because the exact workload is known at kernel generation, we can determine at compile time what  memory operations and computations are necessary. 
In the memory-bound regime, we can load every element of our network parameters and KV cache exactly once. While this requires fully replicating the input, the cost of input data movement is orders of magnintude less than parameter and cache data movement.
\textsc{Cheetah}  automates the index math and static workload distribution and outputs CUDA kernels, thus efficiently generating these static workload schedules. An example of the metaprogamming language, along with a generated kernel, is provided in \Cref{tab:cheetah}, with more advanced examples in \Cref{tab:cheetah_advanced} in the appendix.


As seen in \Cref{fig:partition}, the workload is divided among different thread blocks, then within a thread block into multiple chunks. Finally, each chunk is processed by a series of warps in a series of operations.
Factors like inter-thread block workload partitioning, chunk sizes, and intra chunk workload partitioning are optimized to maximize utilization of hardware resources, especially global-to-shared memory bandwidth.

\begin{figure}
    \centering
    \includegraphics[width=1\linewidth]{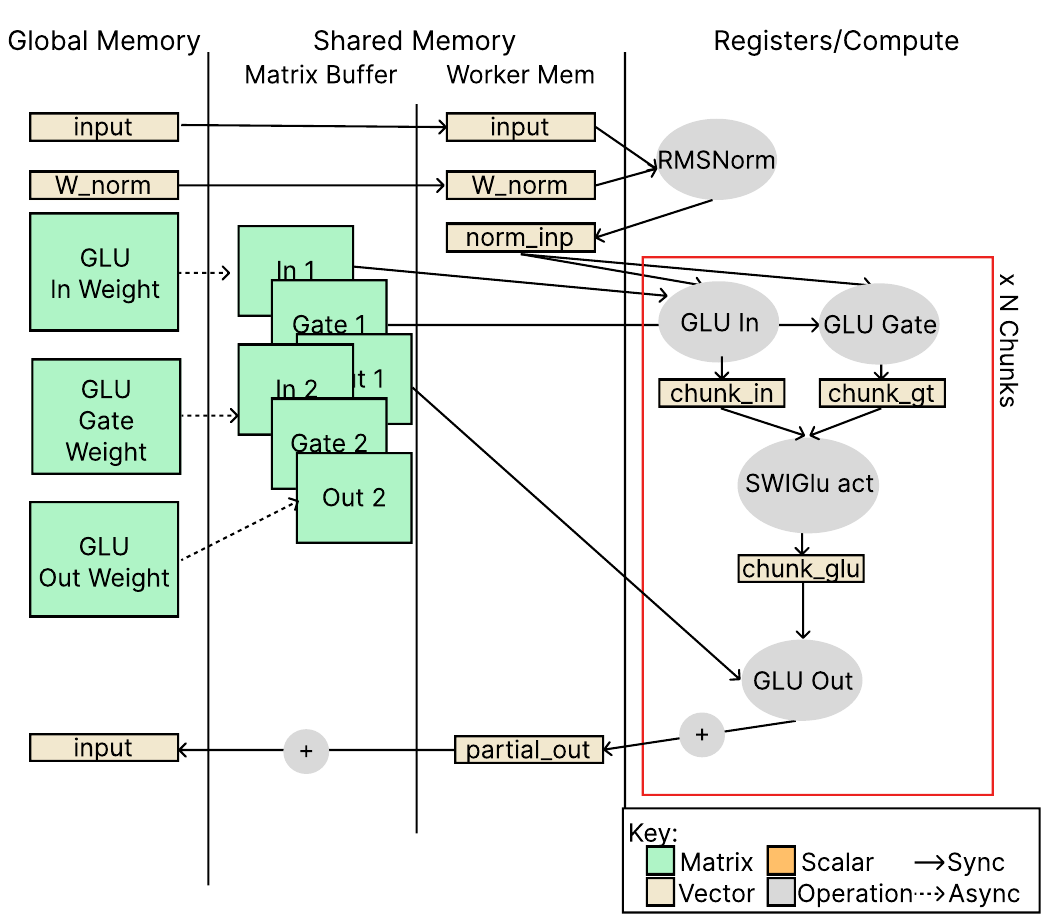}
    \caption{The GLU forward pass. 
    Large matrices (green) are loaded asynchronously from global, and do not wait on consumers. The consumers works synchronously, with no synchronization to global within the sublayer.
    }
    \vspace{-4mm}
    \label{fig:glu}
\end{figure}

\vspace{-1mm}
\paragraph{Memory pipeline.} Each layer has different sized matrices and different shared memory requirements. As shown in \Cref{fig:partition}, we design and implement a unified memory pipeline that allows for the memory movement of arbitrary layers to overlap, and for memory movement to overlap all other operations. This overlapping of memory movement with computation is also referred to as a producer-consumer pipeline design and is used by generally used by fast GEMM-based kernels like FlashAttention-3 for Hopper GPUs \citep{shah2024flashattention3fastaccurateattention}.

Specifically, each matrix listed in \Cref{tab:sublayers} is split into chunks of a fixed size in bytes.  This chunk size remains the same throughout the model, although the data type of the chunk, number of elements, and matrix dimensions are subject to change.
Each matrix's chunks are split among thread blocks at compile time. While all weight matrix chunks are loaded every iteration, KV cache chunks are only loaded as needed.
Due to this structure, different sections of the matrix buffer can store matrix chunks from different layers or sublayers. 
The number of matrix buffers stored at a given time is the ``depth'' of the pipeline, and the size of one matrix buffer is the size of each pipeline stage.

\vspace{-1mm}
\paragraph{Global memory ordering.} One potential issue with our approach is that the inputs for subsequent layers rely on the outputs of prior layer. Given the relaxed memory ordering of CUDA computations, we use a per thread global memory fence coupled with an atomic-based global synchronization mechanism to synchronize volatile reads and writes.
This application of a traditional atomic synchronization ensures all participating threads' global memory operations are visible to all other participating threads before continuing execution to avoid a read-after-write race condition. While atomic-based synchronization is far from novel, using it to combine different transformer layers into a single fused kernel has not been explored before to our knowledge.


\vspace{-1mm}
\subsection{Transformer-specific Kernel Design}
\vspace{-1mm}
We now describe how the above principles are operationalized for the transformer architecture.

\vspace{-1mm}
\paragraph{Attention.}
\label{sub:att}
We divide the attention sublayer into three parts, each of which has synchronous global communication between them: QKV Projection, Core Attention, and Attention Output Projection.
A comparison of standard workload partitioning to ours can be found in \Cref{fig:attention}. \Cref{fig:attn_long} delineates a more detailed diagram of the operations within each of the above parts.

In the batch one case, both QKV Projection and Attention Output projection are matrix vector operations. For these operations, we cache the input in shared memory, and each chunk of the matrix corresponds to some number of rows of the weight matrix.
We compute RMSNorm, RoPE, and then QKV projection, and add the outputs of  Attention Output projection directly to the residual stream.

For Core Attention, the thread blocks are split among the different KV heads, and  
for each KV head, each SM retains a replicated copy of queries which multiply with that particular KV head. The KV cache is then partitioned into matrix chunks, which are fed into the memory buffers. Since our pipelined approach has no memory ordering that guarantees  the elements of the KV cache correspond to the current token, we synchronously load the keys and values for the current token. 

After the entire KV cache has been read and the partial core attention output computed, we  perform a reduction per KV head before writing to a shared buffer. The reduction occurs in three stages. First, each thread block writes required metadata for the \textsc{FlashDecoding}-like reduction, e.g. the maximum logit and the softmax denominator. 
In the second stage, each thread block reads the combined metadata and computes the global softmax denominator. It then rescales its output by the ratio of the thread block's softmax denominator to the combined softmax denominator, as in \citet{milakov2018onlinenormalizercalculationsoftmax}.
Finally, all threads add their rescaled outputs to a cumulative sum in global memory. Compared to the \textsc{FlashDecoding} kernel \citep{flashdecoding}, the reduction is fused into the output computation kernel, removing one activation round trip.

\vspace{-1mm}
\paragraph{GLU.}
\label{sub:glu}
In our kernel, the GLU in/gate operation is a matrix vector operation very similar to the QKV and Attention Output Projection operations in Attention, as shown in \Cref{fig:glu}.
Unlike those operators, the output of each chunk of the  GLU in/gate operation is not written to global memory. Instead, we compute the SwiGLU activation on the outputs of the GLU in/gate function and then multiply it with a matrix chunk of GLU out.
We then keep a vector of size $d_{model}$ for accumulating partial outputs.
With this approach, we can iterate through the entirety of the GLU computation without needing to do any synchronization or reduction.
At the end the computation, we reduce all of these partial outputs along with the input. See \Cref{fig:glu} for a full schematic.

\begin{table*}
\begin{tabular}{p{0.45\textwidth}|p{0.45\textwidth}}
\textbf{Python Implementation} & \textbf{Generated CUDA Kernel} \\
\hline
\begin{lstlisting}[basicstyle=\ttfamily\scriptsize, breaklines=true]
def test_mul2():
    @ch.kernel(ch.Params(n=ty.u32, x=ty.ptr_mut(ty.i32)))
    def kernel(n, x):
        i = ch.block_idx_x() * ch.block_dim_x() + ch.thread_idx_x()
        with ch.if_(i < n):
            x[i] *= 2
    return ch.render(kernel)
\end{lstlisting}
&
\begin{lstlisting}[basicstyle=\ttfamily\scriptsize, breaklines=true]
__global__
void kernel(
    uint32_t n,
    int32_t* x
) {
    uint32_t i = blockIdx.x * blockDim.x + threadIdx.x;
    if (i < n) {
        x[i] = x[i] * 2;
    }
}
\end{lstlisting}
\end{tabular}
\caption{Upper Left: Python kernel source. Upper Right: Generated CUDA kernel. Our metaprogramming language generates CUDA kernels from a Python interface. We use this to automate index math, perform static workload scheduling, and build kernel metaprogramming tools. More advanced examples can be found  in \Cref{tab:batched} in the appendix.}

\label{tab:cheetah}
\end{table*}

%% file: sections/results.tex
\begin{table}[t]
    \centering
    \begin{tabular}{llccc}
        \toprule
         \textbf{Model} &  \textbf{Prefill}  &\multicolumn{3}{c}{\textbf{Tokens per second}} \\
         \textbf{Size} & \textbf{Size} &  GPTFast & vLLM  & \flashformer \\
         \midrule
        8B & 128 & 168 & 146   &184  \\
(\texttt{BF16})    & 1024 & 159 & 144   &182 \\
        & 3072 & 158 &  141  &180 \\
        & 6144 & 143 & 138  &176 \\
         \midrule
8B   &               128 & 290 &  193 & 311 \\
(\texttt{INT4})          & 1024 & 284 & 186 & 310 \\
        & 3072 & 266 & 189 & 307 \\
        & 6144 & 228 & 175 & 301\\
        \midrule
70B        &          128 & 47 & - & 48\\
(\texttt{INT4})   &          1024 & 45 & - & 48\\
&          3072 & 43 & - & 47\\
&          6144 & 40 & - & 45\\
\bottomrule        
    \end{tabular}
    \vspace{-2mm}
    \caption{Prefill size and tokens per second (TPS) for the model configurations in the batch one case for the LLama 3.1 series of models. We were unable to get vLLM numbers for LLama 3.1 70B quantized results due to system errors. }
    \vspace{-4mm}    
    \label{tab:llama 3 8 unquantized TPS}
\end{table}

\begin{figure}[t]
    \vspace{-2mm}
    \centering
    \includegraphics[width=0.9\linewidth]{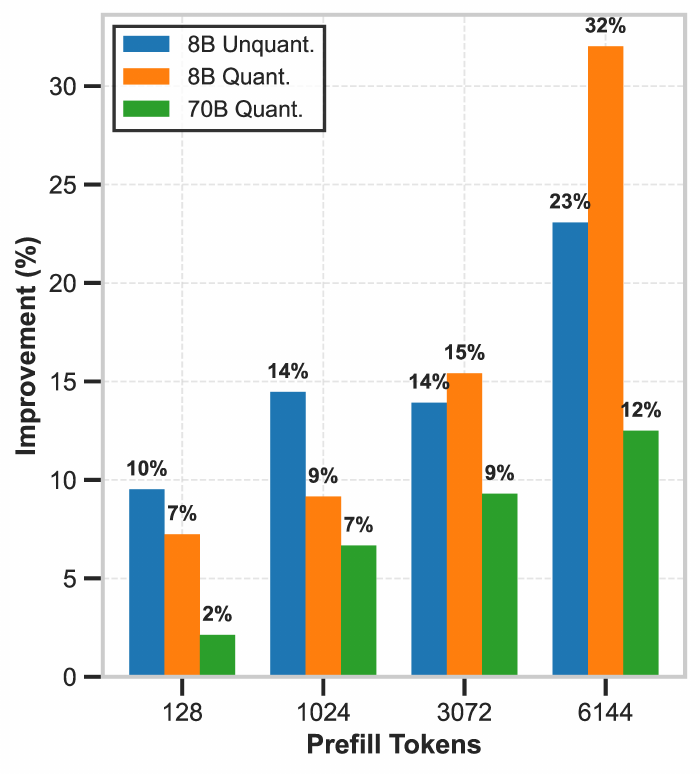}
        \vspace{-4mm}
    \caption{Percentage improvements in decoding tokens per second (TPS) of our kernel over GPTFast.}
    \label{fig:results}
    \vspace{-4mm}
\end{figure}

\vspace{-2mm}
\section{Experiments}
\vspace{-1mm}

\subsection{Experimental Setup}
\vspace{-1mm}
We apply our kernel to three models: Llama 3.1 8B unquantized, Llama 3.1 8B quantized, and Llama 3.1 70B quantized. We compare with two strong baselines that are widely used in practice: GPTFast\footnote{\url{https://github.com/pytorch-labs/gpt-fast}} and vLLM\footnote{\url{https://github.com/vllm-project/vllm}} \citep{10.1145/3600006.3613165}. 
We enable all optimizations where possible. 
In particular, we enable CUDA graphs to decrease kernel launch overheads in our baselines.

We test a variety of prefill lengths, and generate 128 tokens per output. We only measure decoding speed, as prefill is compute bound. As a result, we only need to generate enough tokens to achieve a good estimate of the per-token latency with a given amount of prefill. All experiments are performed on one H100 SXM GPU. 

\vspace{-1mm}
\subsection{Main Results}
\vspace{-1mm}
\paragraph{Batch size = 1 case.} We first focus on the batch size = 1 setting and show the results in \Cref{tab:llama 3 8 unquantized TPS} and \Cref{fig:results}.
On unquantized Llama 3.1 8B, we find that our kernel achieves a 8-20\% speedup over GPTFast, and up to a 61\% speedup over vLLM. We observe similar improvements with quantization, where we use  GPTQ \citep{frantar-gptq} to perform weight-only integer quantization to 4 bits   with a group size of 128. 
While this computation is likely bottlenecked by the operations needed for dequantization and not memory bandwidth, our approach allows us better to overlap dequantization with memory movement and computation.  

We also evaluate on a large quantized model, Llama 3.1 70B. 
 Our expectation was that the gains at 70B would be smaller than at 8B.
 This is because larger models have more opportunities for overlapping within a kernel. This decreases the need for overlapping between kernel launches. 
 The results, shown in \Cref{tab:llama 3 8 unquantized TPS}, are small improvements (2\%) at short sequence lengths increasing to 13\% at longer sequence lengths.

\vspace{-1mm}
\paragraph{Larger batch sizes.} We perform experiments with batch size two and four with the unquantized 8B model. 
We consider the case where KV caches are entirely distinct between elements of the batch, and generation for all queries within the batch starts and stops at the same time.
These batched kernels are still input stationary: weights are read in exactly once, and inputs are fully replicated on each SM. 
The synchronization and bandwidth savings from the batch size one case hold in the slightly larger batch setting, as shown in \Cref{tab:batched}. However, memory bandwidth utilization decreases, as the increased shared memory pressure does force smaller memory pipelines or a minimal L1 cache.

\vspace{-1mm}
\paragraph{Discussion.} Two trends emerged in our experiments.
First, \flashformer had the largest speedups at the longest sequence lengths.
We suspect this is due to our method of partitioning the KV cache. The size of our pipeline stages is 64KB, which corresponds to 256 keys or values with head dim 128 and \texttt{BF16} precision. 
With 8 SMs per KV head, we need 4096 keys and values to provide one matrix chunk per participating SM.
At longer sequence lengths this problem vanishes. Instead, \flashformer overlaps more of the core attention computation and reduction with memory movement, improving bandwidth utilization.
Future work can explore better optimization of  workloads at different sequence lengths.

The second trend we observe is that smaller models benefit more from our technique than larger models, as evidenced by \Cref{fig:results}.
This is likely due to the fact that very large models have more opportunities for overlapping within each operation, and thus gain less benefit from overlapping between operations.

\begin{table}[t]
\centering
\begin{tabular}{c c c c c}
    \toprule
    Batch Size & SL & GPTFast & vLLM & \flashformer \\
    \midrule
     2&  128 & 308 & 286 & 328\\
     4 & 128 & 608 & 564 & 616\\
     \midrule
     2&  1024 & 556 & 560 & 648\\
     4 & 1024 & 504 & 548 & 600\\
     \bottomrule
\end{tabular}
\caption{Batched tokens per second for unquantized Llama 3.1 8B under various settings. SL Stands for sequence length.}
\vspace{-2mm}
\label{tab:batched}
\end{table}

\vspace{-1mm}
\subsection{Ablations \& Analysis}
\vspace{-1mm}
We next perform a series of ablations and analyses.
\vspace{-1mm}
\paragraph{Pipeline depth.}
We investigate how increasing pipeline stages affects kernel performance for the unquantized Llama 3.1 8B model. Our kernel design requires a minimum of two concurrent memory buffers during GLU computation. Consequently, a two-stage pipeline cannot overlap memory operations during GLU in/gate operations. 
The kernel was primarily optimized for a 64KB pipeline stage size. 
Performance measurements comparing two versus three pipeline stages at 3072 sequence length reveal that three stages achieves a 5\% speedup over two stages, as shown in \Cref{tab:pipeline_depth}.

Due to shared memory capacity constraints, we cannot implement four pipeline stages at 64KB per stage. To test deeper pipelines, we halved the pipeline stage size, though these configurations were not as thoroughly optimized as the larger stage implementations. Results indicate that while four pipeline stages outperform three, five stages do not yield further improvements over four. At this depth, pipeline management overhead negates the benefits of deeper pipelining.

\begin{table}[t]
    \centering
    \begin{tabular}{ccc}
    \toprule
         Pipeline Stage Size & Number of Stages & TPS \\
     \midrule
         64KB & 2 & 170\\
         64KB & 3 & 179\\
    \midrule 
         32KB & 3 & 143\\
         32KB & 4 & 138\\
         32KB & 5 & 130\\
         
    \bottomrule
    \end{tabular}
    \caption{Llama 3.1 8B unquantized pipeline depth ablation at sequence length 3072. We primarily optimize for 64KB pipeline stage size, and thus have worse workload partitioning at 32KB.}
    \label{tab:pipeline_depth}
\vspace{-2mm}
\end{table}

\begin{figure}
    \centering

    \includegraphics[width=\linewidth]{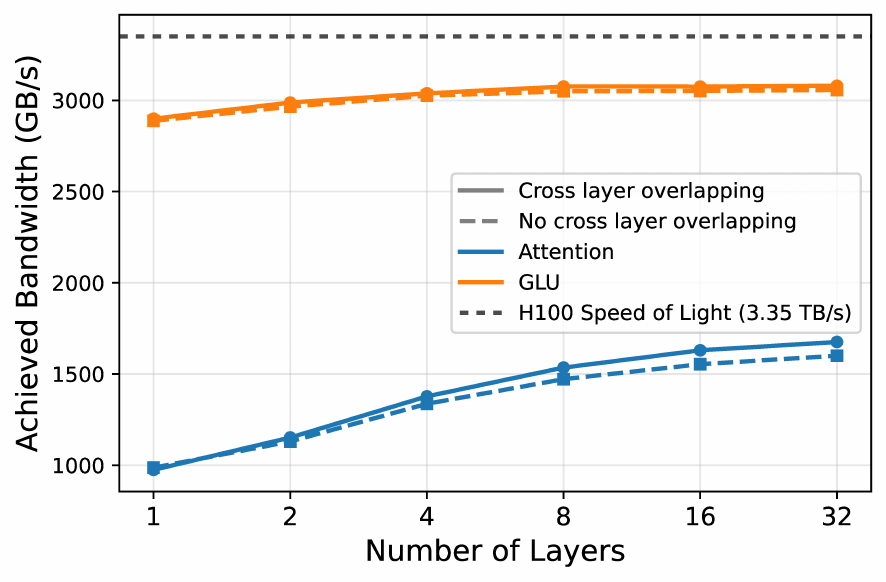}
    \caption{Attention and GLU memory bandwidth in an unquantized Llama 3.1 8B kernel. While GLU nears hardware peak, our attention implementation achieves closer to 50\% of peak bandwidth. As GLU dominates memory usage, net bandwidth is much closer to GLU memory bandwidth.}
    \vspace{-6mm}
    \label{fig:attn_glu_comparison}
\end{figure}

\vspace{-2mm}
\paragraph{Attention \& GLU kernels.} We measure the latency and  memory bandwidth of stacked layers of Attention and GLU blocks, similar to our motivating study with stacked linear layers.  The results are in \Cref{tab:latency} and \Cref{fig:attn_glu_comparison}. 

For latency, we see that double digit latency improvements occur from cross layer fusion, and single digit improvements occur from cross layer memory overlapping, regardless of layer type. Additionally, we ablate the width and depth of our pipeline for the Attention and GLU kernels. While the the attention and linear layers are fastest with a depth 2 pipeline, the GLU layer is fastest with a depth 3 pipeline. As GLU represents the majority of parameters and latency, the composite kernel is fastest with a depth 3 pipeline. Future work may look at adaptively changing pipeline size throughout the model

For memory bandwidth, while stacked GLU blocks are able exceed 90\% of machine peak,  our stacked attention blocks achieve roughly 50\% of machine peak. Compared to GLU layers, attention layers have more logic, special functions, and synchronization with less memory movement to hide them. Note that as nearly eighty percent of layer parameters are in GLU layers, the bandwidth of our resulting kernel is much closer to the bandwidth of the stacked GLU layer.





\begin{table}[]
\centering
\begin{tabular}{l c c c c}
\toprule
\textbf{Module} & \textbf{Layers} & \textbf{Baseline} & \textbf{Fused} & \textbf{+Overlap} \\
\midrule
 & 1 & 0.089 & 0.089 & 0.090 \\
Attn & 4 & 0.357 & 0.264 & 0.256 \\
 & 32 & 2.857 & 1.761 & 1.683 \\
\midrule
 & 1 & 0.122 & 0.122 & 0.122 \\
GLU & 4 & 0.488 & 0.466 & 0.464 \\
 & 32 & 3.905 & 3.689 & 3.661 \\
\bottomrule
\end{tabular}
\vspace{-2mm}
\caption{Component-wise latencies for multilayer attention and GLU kernels, in ms. Baseline does not fuse any layers. Fused  refers to a  kernel that fuses across layers without memory overlapping. + Overlap incorporates cross-layer memory overlapping.}
\label{tab:latency}
\vspace{-2mm}
\end{table}

\vspace{-1mm}
\section{Discussion \& Limitations}
\vspace{-1mm}
\subsection{Handwritten vs Compiled Kernels}
One interesting direction of research is automatically compiling whole-model kernels. \textsc{Mirage} \citep{wu2024mirage} recently added support to user their kernel generator to automatically generate whole model kernels. We implemented an optimized Llama3 in for Hopper GPUs in their framework and compared TPS numbers. Unfortunately their benchmarking tools combine prefill and decoding latency, but even when generating without any prefill the peak performance we achieved was 153 TPS with LLama3 8B. The difference in performance appears to be due to better work partitioning, decreased synchronization, and greater overlapping in our kernel. Currently, the greater exploration space opened up by handwriting kernels leads to ~20\% performance gains in our case.

\vspace{-1mm}
\subsection{Compile vs. Runtime optimization}
\vspace{-1mm}
The work partitioning within our kernels---in particular, the partitioning of matrix chunks between thread blocks---is determined at compile time. 
This  simplifies kernel logic and minimizes communication to global memory. 

However, there are drawbacks to this strategy. 
At every point of synchronization, the entire GPU is held back by the slowest SM. 
Furthermore, it can lead to inefficiency when the amount of work changes at runtime, which happens with the KV cache.
For our workloads, we tried to avoid this problem through careful division of memory loads.
Works such as ThunderMLA \citep{thundermla} also implement a dynamic workload partitioning for a persistent \textsc{FlashAttention} style Multi Latent Attention \citep{deepseekv2} kernel, controlled through a global instruction buffer.
Their persistent kernel accesses a global memory buffer to determine what problem to work on next. 

We hope that future work explores dynamic runtime optimized versions of \flashformer, and explores the tradeoff between increased inter thread block communication to better balance the workload and decreased inter thread block communication to minimize per thread block latency.

\subsection{Limitations}

Our work is limited insofar as we target the single-GPU, low-batch setting for the Hopper family of architectures. We focus on this setting to demonstrate the practicality of the core ideas introduced in this paper, including aggressive specialization and the feasibility of a kernel design with a static workload schedule. 
We hope that future work applies these ideas to larger-batch, multi-GPU settings on different families or architectures, as well as to non-transformer models.
Furthermore while we contribute a custom kernel metaprogramming library with customizable kernel templates, further work is needed to extend our abstractions to a greater variety of models and  multi-GPU settings.

%% file: sections/RW.tex
\section{Related Work}

\paragraph{Specialized Transformer model kernels.}
Due to the fact that the linear / GLU layers are standard linear algebra operations, most of the focus of specialized transformer kernel design has focused on optimizing attention.
The FlashAttention \citep{dao2022flashattention,dao2023flashattention,shah_flashattention-3_2024} 
restructures the attention computation to a numerically equivalent version that fused all of core attention.
This general strategy is widespread and was used in many attention or attention-like operations.

Accelerator Kernel DSLs, such as Triton~ \citep{10.1145/3315508.3329973}, are used to simplify the kernel generation process by grouping computation into conceptual tiles. Specialized tools such as 
FlexAttention~\citep{guessous2024flexattention} 
further allow for simple kernel generation for attention like workloads.
Recently, automatic fusion of pointwise operators has continued to improve and become a staple of all modern machine learning libraries. XLA~\citep{50530} and PyTorch Compile~\citep{Ansel_PyTorch_2_Faster_2024} automate fusion of pointwise operations for Jax~\citep{jax2018github} and PyTorch respectively.

Concurrent to this work, ThunderMLA \citep{thundermla} introduce a persistent kernel for MultiHead Latent Attention ~\citep{deepseekv2} layers that has each thread block dynamically schedule its own workload. 

\vspace{-1mm}
\paragraph{Low batch inference optimizations.}
Better kernel design is far from the only research area pursuing greater inference efficiencies.
Two of the most important research directions in low batch inference are speculative decoding and quantization.

Speculative decoding \citep{leviathan2022fast} performs autoregressive inference on a small model and multi-token inference on a larger model. This relies on similarity between the model distributions and the relative speed of the small model.
The critical idea was that the large verifier model would run at an effective batch size as long as the number of consecutive tokens fed into the model. Conditioned on the tokens verification, a single large model load could generate multiple tokens.
Further iterations \citep{10.5555/3692070.3692273, ankner2024hydra,li2024eagle} improved upon this formula by using draft heads computed on the model's outputs instead of a separate draft model.

Weight quantization methods \citep{frantar-gptq, yao2022zeroquant,chee2023quip} have accelerated decoding by enabling lower and lower bit precisions for weight loads.
This decreases the cost of memory movement, and thus improves inference speed.
However, this is not perfect, as most of these methods utilize scarce GPU resources for dequantization.

%% file: sections/conclusion.tex
\section{Conclusion}

We introduce a new method for overlapping operations between layers for transformer decoding workloads.
We develop a family of kernels that leverage this new method of overlapping, and 
empirically verify the performance of these models on real workloads.

\section*{Acknowledgements}
This work was  supported by the National Science Foundation under CAREER Award No. 2441872 and the MIT-IBM Watson AI Lab.

%% file: appendices/algorithims.tex
\newpage

\section{Metaprogramming Language Example}
\Cref{tab:cheetah_advanced} shows more examples from our \textsc{Cheetah} metaprogramming language.
\label{sec:appdx_cheetah}
\begin{table*}[t!]
\begin{tabular}{p{0.45\textwidth}|p{0.45\textwidth}}
\textbf{Python Implementation} & \textbf{Generated CUDA Kernel} \\
\centering
\scriptsize

\begin{lstlisting}[basicstyle=\ttfamily\scriptsize, breaklines=true]
class AsyncBarrierGTSCopyFn(SmartScopeFn):
...
def setup(
    self,
    src_cfg: SDIPConfig,
    dst_cfg: SDIPConfig,
    thread_idx: DimName,
    cpy_idx: DimName,
):
...
def generate(
    self,
    ix: Indices,
    src_ptr: SafeDataIndexPtr,
    dst_ptr: SafeDataIndexPtr,
    barrier: SharedMBarrier,
):

    size = (ix.size(self.cpy_idx) *
    dst_ptr.get_elem_size())
    ix.set_index(self.cpy_idx, ch.const(0, ty.i32))
    assert size % 16 == 0
    with ch.if_(ix[self.thread_idx] == 0):
        ch.asm(
            f"cp.async.bulk.shared::cluster.global.mbarrier::complete_tx::bytes [$dst_ptr], [$src_ptr], {size}, [$barrier];",
            dst_ptr=dst_ptr.idx_offset(),
            src_ptr=src_ptr.idx_offset(),
            barrier=barrier.ptr,
        )
        ch.asm(
            f"mbarrier.expect_tx.relaxed.cta.shared.b64 [$barrier], {size};",
            barrier=barrier.ptr,
        )
    ch.raw_stmt("__syncwarp();")
...
async_gts_copy_fn(ix, src_ptr, shared_ptr, shared_barrier)
\end{lstlisting}
&
\begin{lstlisting}[basicstyle=\ttfamily\scriptsize, breaklines=true]
int32_t elem_idx0 = 0;
if (thread_idx == 0) {
    asm ("cp.async.bulk.shared::cluster.global.mbarrier::complete_tx::bytes [%0], [%1], 32768, [%2];" : : "l"(&shared[elem_idx0]), "l"(&x[elem_idx0]), "r"(mbar_shmem));
    asm ("mbarrier.expect_tx.relaxed.cta.shared.b64 [%0], 32768;" : : "r"(mbar_shmem));
}
__syncwarp();
\end{lstlisting}
\\
\midrule
\begin{lstlisting}[basicstyle=\ttfamily\scriptsize, breaklines=true]
self.qk_compute_fn(
    ix,
    self.shared_query_buffer_ptr,
    k_buffer,
    self.scores_ptr,
    self.mem_pipeline.consumer_barrier,
)
\end{lstlisting}
&
\begin{lstlisting}[basicstyle=\ttfamily\scriptsize, breaklines=true]
...
for (int32_t k_iter_idx1 = 0; k_iter_idx1 < 16; k_iter_idx1 += 1) {
    for (int32_t n_iter_idx1 = 0; n_iter_idx1 < 4; n_iter_idx1 += 1) {
        ...
        asm volatile (
            "mma.sync.aligned.m16n8k8.row.col.f32.bf16.bf16.f32 { %0, %1, %2, %3}\n"
            ", { %8, 0.0}\n"
            ", { %9}\n"
            ", { %4, %6, %5, %7};\n" : "=f"(f0), "=f"(f1), "=f"(f2), "=f"(f3) : "f"(out_cache0[n_iter_idx2 * 2 + 0]), "f"(0.0f), "f"(out_cache0[n_iter_idx2 * 2 + 1]), "f"(0.0f), "r"(reinterpret_cast<int32_t*>(&mat1_cache0[k_iter_idx1 * 2 + 0])[0]), "r"(reinterpret_cast<int32_t*>(&mat2_cache0[n_iter_idx2 * 2 + 0])[0]));
        out_cache0[n_iter_idx2 * 2 + 0] = f0;
        out_cache0[n_iter_idx2 * 2 + 1] = f1;
    }
}
...
\end{lstlisting}
\end{tabular}
\vspace{2mm}
\caption{
Upper Left: A \textsc{Cheetah} Python class that used for bulk async copies. Through the use of input configs and knowledge of valid use of the assembly instruction, it automates validation . When called during kernel generation it first checks the validity of the barrier and copy size, then solves index math, then outputs the code snippet on the right. Names can be specified for CUDA variables, and assembly and raw CUDA can be inserted at any point of program flow.
Lower Left: an example of a higher level \textsc{Cheetah} function, for the QK projection in attention. After the $m$,$k$, and $n$ of the matmul are specified, loops and assembly instructions are automatically generated. In this particular example, only half of the outputs of the MMA are used due to the size of the computation, which is also automatically handled.}
\label{tab:cheetah_advanced}
\end{table*}